# Scientific Discovery by Machine Intelligence: A New Avenue for Drug Research

Carlo A. Trugenberger

InfoCodex AG, Semantic Technologies, Bahnhofstrasse 50, Buchs, CH-9470 Switzerland

Email: c.trugenberger@InfoCodex.com

## Abstract

The majority of big data is unstructured and of this majority the largest chunk is text. While data mining techniques are well developed and standardized for structured, numerical data, the realm of unstructured data is still largely unexplored. The general focus lies on "information extraction", which attempts to retrieve known information from text. The "Holy Grail", however is "knowledge discovery", where machines are expected to unearth entirely new facts and relations that were not previously known by any human expert. Indeed, understanding the meaning of text is often considered as one of the main characteristics of human intelligence.

The ultimate goal of semantic artificial intelligence is to devise software that can "understand" the meaning of free text, at least in the practical sense of providing new, actionable information condensed out of a body of documents. As a stepping stone on the road to this vision I will introduce a totally new approach to drug research, namely that of identifying relevant information by employing a self-organizing semantic engine to text mine large repositories of biomedical research papers, a technique pioneered by Merck with the InfoCodex software. I will describe the methodology and a first successful experiment for the discovery of new biomarkers and phenotypes for diabetes and obesity on the basis of PubMed abstracts, public clinical trials and Merck internal documents. The reported approach shows much promise and has potential to impact fundamentally pharmaceutical research as a way to shorten time-to-market of novel drugs, and for early recognition of dead ends.

# Big data: challenges and opportunities

Rivers of ink have been poured to describe the data deluge that is increasingly defining our information society. While I do not want to dwell too long on something we all are experiencing daily, the concrete numbers are nonetheless staggering [1]. Here are some examples:

- In 2007 more data have been accumulated than can fit on all of the world's available storage.
- In 2011 this number has reached the limit of twice as much data as can be stored on all of the world's storage i.e. 1200 billions gigabytes.
- The CMS detector at the CERN LHC accelerator accumulates data at a rate of 320 terabits/s, which makes it necessary to filter data by hardware "on the way" to reduce to flux to "only" 800 Gbp/s.
- Wal-Mart feeds 1 million customer transaction/hour into its databases .
- Internet: 1 trillion unique URLs have been indexed by Google.
- 12.8 million blogs have been recently recorded, not counting Asia, this number is growing exponentially.
- The number of emails sent per day in 2010 was 294 billion.
- In 2008 Google received 85'000 CVs for the one single post of software engineer.

These numbers pose huge challenges on both hardware and software. However, as is usually the case, challenges and opportunities go hand in hand. In this paper I shall concentrate on the opportunity side of the equation.

Data come in two flavours: structured and unstructured. Structured data consist typically of numbers organized in structures, like tables, charts or series. Unstructured data are essentially everything else and make up around 85% [2] of the data deluge. Of these, the vast majority is text, the rest being pictures, video and sound tracks. In this paper I shall concentrate on text data.

There is only one thing you can do with numbers: analyze them to discover relationships and dependencies. The basic method to do this is statistical analysis, whose development was initiated in the $17^{th}$ century with the works of Pascal, Fermat, de Moivre, Laplace and Legendre and got new impetus in the late $19^{th}$ and early $20^{th}$ century from Sir Francis Galton and Karl Pearson [3]. Today, statistical analysis if often complemented by methods from computer science and information theory to detect unsuspected patterns and anomalies in very large databases, a technique that goes under the name of data mining [4].

While statistical analysis and data mining are complex and require trained specialists, unstructured data pose even bigger challenges. First of all there are two things you can do with text: teach machines to understand what the text in a given document means and have them "read" large quantities of text documents to uncover hidden, previously unnoticed correlations pointing to entire new knowledge. Both are very difficult but the latter is far more difficult than the former.

# Information extraction and knowledge discovery in research papers

Understanding written language is a key component of human intelligence. Correspondingly, doing something useful with large quantities of text documents that are out of reach for human analysis requires, unavoidably some form of *artificial intelligence* [5]. This is why handling unstructured data is harder than analyzing their numerical counterpart, for which well-defined and developed mathematical methods are readily available. Indeed, there is as yet no standard approach to text mining, the unstructured counterpart to data mining.

There are several approaches to teach a machine to comprehend text [6-8]. The vast bulk of research and applications focuses on natural language processing (NLP) techniques for *information extraction* (IE). Information extraction aims to identify mentions of named entities (e.g. "genes" in life science applications) and relationships between these entities (as in "is a" or "is caused by"). Entities and their relations are often called "triples" and databases of identified triples "triple stores". Such triple stores are the basis of the Web 3.0 vision, in which machines will be able to automatically recognize the meaning of online documents and, correspondingly, interact intelligently with human end users. IE techniques are also the main tool used to curate domain-specific terminologies and ontologies extracted from large document corpora.

Information extraction, however, is not thought for *discovery*. By its very design, it is limited to identifying semantic relationships that are explicitly lexicalized in a document: by definition these relations are known to the human expert who formulated them. The "Holy Grail" [9] of the text mining, instead is *knowledge discovery* from large corpora of text. Here one expects machines to generate novel hypotheses by uncovering previously unnoticed correlations from information distributed over very large pools of documents. These hypotheses must then be tested experimentally. Knowledge discovery is about unearthing implicit information versus the explicit relations recovered by information extraction. The present paper is about machine knowledge discovery in the biomedical and pharmacogenomics literature.

# 21$^{st}$ century challenges for pharmaceutical research

Pharmaceutical research is undergoing a profound change. The deluge of molecular data and the advent of computational approaches to analyze them have revolutionized the traditional process of discovering drugs by happenstance in natural products or synthetizing and screening large libraries of small molecule compounds. Today, computational methods permeate so many aspects of pharmaceutical research that one can say that drugs are "designed" rather than "discovered" [10,11].

Molecular data found in genomics and proteomics databases are typically structured data. As in other domains, the bulk of the computational effort in the pharmaceutical industry goes into crunching structured molecular data. There is, however another, even larger source of valuable information that can potentially be tapped for discoveries: repositories of research documents. One of the best known of these repositories, PubMed, contains already more than

20 millions citations and these are growing at a once inconceivable rate of almost 2 papers/minute [12].

The value of the information in these repositories of research is huge. Each paper by itself constitutes typically a very focused study on one particular biomedical subject that can be easily comprehended by other experts in the same field. It is to be expected, however that there are also far-reaching correlations between the results of different papers or different groups of papers. Uncovering such hidden correlations by hand borders on the impossible since, first, the quantity of such papers is by now far beyond the reach of human analysis and, secondly, the expertise to understand papers in different areas of research is very hard to find in the same individual in today's era of ever increasing specialization. The potential competitive advantage for the first companies to succeed in the task of discovering new scientific knowledge this way is considerable, both in speeding up research and in cutting costs. This is why machine knowledge discovery, if successful, has the potential to revolutionize pharmaceutical research. Not only could one test hypotheses in silico but the actual generation of these hypotheses would be in silico, with obvious disruptive advantages.

## Discovering biomarkers and phenotypes by text mining?

To explore if this vision of a new way to generate scientific discovery by machine intelligence is feasible, Merck, in collaboration with Thomson Reuters, devised a pilot experiment in which the InfoCodex semantic engine was used for the specific and concrete task to discover unknown/novel biomarkers and phenotypes for diabetes and/or obesity (D&O) by text mining diverse and numerous biomedical research texts [13]. Here I will summarize the key points of the methods and the main results.

The choice fell on biomarkers and phenotypes since these play a paramount role in modern medicine. Drugs of the future will be targeted to populations and groups of individuals with common biological characteristics predictive of drug efficacy and/or toxicity. This practice is called "individualized medicine" or "personalized medicine" [10]. The revealing features are called "biomarkers" and "phenotypes".

A biomarker is a characteristic that is objectively measured and evaluated as an indicator of normal biologic processes, pathogenic processes, or pharmacologic responses to a therapeutic intervention. In other words, a biomarker is any biological or biochemical entity or signal that is predictive, prognostic, or indicative of another entity, in this case, diabetes and/or obesity.

A phenotype is an anatomical, physiological and behavioral characteristic observed as an identifiable structure or functional attribute of an organism. Phenotypes are important because phenotype-specific proteins are relevant targets in basic pharmaceutical research.

Biomarkers and phenotypes constitute one of the "hot threads" of diagnostic and drug development in pharmaceutical and biomedical research, with applications in early disease identification, identification of potential drug targets, prediction of the response of patients to medications, help in accelerating clinical trials and personalized medicine. The biomarker market generated $13.6 billion in 2011 and is expected to grow to $25 billion by 2016 [14].

The object of the experiment was for the InfoCodex semantic engine to *discover unknown/novel biomarkers and phenotypes for diabetes and/or obesity (D&O)* by text mining a diverse and sizable corpus of unstructured, free-text biomedical research documents constituted by:

- PubMed [15] abstracts with titles: 115,273 documents
- Clinical Trials [16] summaries: 8,960 summaries
- Internal Merck research documents, about one page in length: 500 documents.

The output D&O related biomarkers and phenotypes proposed by the machine were then compared with Merck internal and external vocabularies/databases including UMLS [17], GenBank [18], Gene Ontology [19], OMIM [20], and the Thomson Reuters [21] D&O biomarker databases. By design, the experiment was handled strictly as a "blind experiment": no expert input about D&O biomarkers/phenotypes was provided and no feedback from preliminary results was used to improve the machine-generated results.

## The InfoCodex semantic engine

InfoCodex is a semantic machine intelligence software designed specifically to analyze very large document collections as a whole and thereby unearth associative, implicit and lexically unspecified relationships. It does so by unsupervised semantic clustering and matching of multi-lingual documents. Its technology is based on a combination of an embedded universal knowledge repository (the InfoCodex Linguistic Database, ILD), statistical analysis and information theory [22], and self-organizing neural networks (SOM) [23].

The ILD contains multi-lingual entries (words/phrases) collected into cross-lingual synonym groups (semantic clouds) and systematically linked to a hypernym (taxon) in a universal 7-level taxonomy. With its almost 4 million classified entries, the ILD corresponds to a very large multi-lingual thesaurus (for comparison, the *Historical Thesaurus of the English Oxford Dictionary*, often considered the largest in the world, has 920,000 entries).

Information theory and statistics [22] are used to establish a 100-dimensional content space defined on the ILD that describes the documents in an optimal way. Documents are then modeled as 100-dimensional vectors in this optimal semantic space. Information-theoretic concepts such as entropy and mutual entropy are used together with the ILD to disambiguate the meaning of polysemous words based both on the document-specific context and the collection-wide environment.

Finally, the fully automatic, unsupervised categorization on the optimal semantic space is achieved by a proprietary variant of Kohonen's self-organizing map [23]. In particular, prior to starting the unsupervised learning procedure, a coarse group rebalancing technique is used to construct a reliable initial guess for the SOM. This is a generalization of coarse mesh rebalancing [24] to general iterative procedures, with no reference to spatial equations as in the original application to neutron diffusion and general transport theory in finite element analysis. This procedure considerably accelerates the iteration process and minimizes the risk of getting stuck in a sub-optimal configuration. The SOM creates a thematic landscape according to and optimized for the thematic volume of the entire document collection.

Essentially, the combination of the embedded ILD with the self-organized categorization on an automatically determined optimal semantic space correspond to a dynamic ontology, in which vertical "is-a" relations are encoded and horizontal relations like "is-correlated-with" are determined dynamically depending on content.

For the comparison of the content of different documents with each other and with queries, a similarity measure is used which is composed of the scalar product of the document vectors in the 100-dimensional semantic space, the reciprocal Kullback–Leibler distance [25] from the main topics, and the weighted score-sum of common synonyms, common hypernyms and common nodes on higher taxonomy levels.

As a final result, a document collection is grouped into a two-dimensional array of neurons called an information map. Each neuron corresponds to a semantic class; i.e., documents assigned to the same class are semantically similar. The classes are arranged in such a way that the thematically similar classes are nearby (Figure 1).

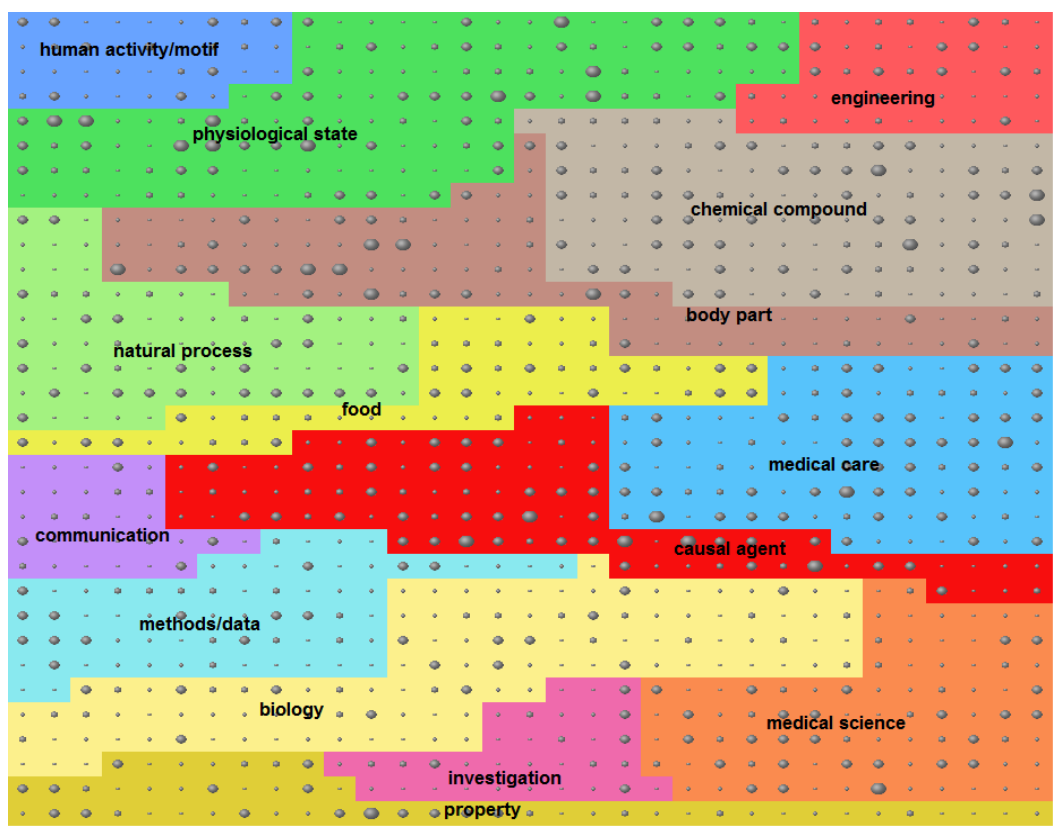

**Figure 1 InfoCodex information map. InfoCodex information map obtained for the approximately 115,000 documents of the PubMed repository used for the present experiment.** The size of the dots in the center of each class indicate the number of documents assigned to it.

The described InfoCodex algorithm is able to categorize unstructured information. In a recent benchmark, testing the classification of multi-lingual, "noisy" Web pages, InfoCodex reached the high clustering accuracy score F1 = 88% [26]. Moreover, it extracts relevant facts not only from single documents at hand, but it considers document collections as a whole and identifies dispersed and seemingly unrelated facts and relationships like assembling the scattered pieces of a puzzle.

# Text mining with InfoCodex in search of new biomarkers/phenotypes

The text mining procedure involved four steps:

**Generation of reference models:** in this step the software had to determine the meaning of the concept "biomarker/phenotype for D&O". Since no input by human experts was allowed in the experiment, the only way to do this was by a generic literature search via the autonomous InfoCodex spider agents: 224 reference biomarkers/phenotypes were found. The documents containing these reference terms were then clustered by InfoCodex and for each group a representative feature vector in the optimal semantic space was established. These feature vectors constitute mathematical models on semantic space of what, e.g. "biomarker for diabetes" means.

**Determination of the meaning of unknown terms:** the ILD contained at the time of the experiment about 20,000 genes and proteins (up to around 100'000 presently). Nonetheless it was not guaranteed to identify all possibly relevant candidates by a simple database look-up. Fortunately, the architecture of InfoCodex allows to infer the meaning of unknown terms by combining its "hard-wired" internal knowledge base with the association power of neural networks. Some examples of the meanings inferred by InfoCodex are presented in Table 1.

Table 1: InfoCodex computed meanings

| *Unknown Term* | *Constructed Hypernym* | *Associated Descriptor 1* |
|---|---|---|
| Nn1250 | clinical study | insuline glargine |
| Tolterodine | cavity | overactive bladder |
| Ranibizumab | drug | macular edema |
| Nn5401 | clinical study | insulin aspart |
| Duloxetine | antidepressant | personal physician |
| Endocannabinoid | receptor | enzyme |
| Becaplermin | pathology | ulcer |
| Candesartan | cardiovascular disease | high blood pressure |
| Srt2104 | medicine | placebo |
| Olmesartan | cardiovascular medicine | amlodipine |
| Hctz | diuretic drug | hydrochlorothiazide |
| Eslicarbazepine | anti nervous | Zebinix |
| Zonisamide | anti nervous | Topiramate Capsules |
| Mk0431 | antidiabetic | sitagliptin |
| Ziprasidone | tranquilizer | major tranquilizer |
| Psicofarmcolagia | motivation | incentive |
| Medoxomil | cardiovascular medicine | amlodipine |

InfoCodex computed meanings of some unknown terms from the experimental PubMed collection.

The meaning of unknown terms is estimated fully automatically; i.e., no human intervention was necessary and no context-specific vocabularies had to be provided as in most related approaches [27]. The meaning had to be inferred by the semantic engine only based on machine intelligence and its internal generic knowledge base, and this automatism is one of the main innovations of the presented approach. Some of the estimated hypernyms are completely correct: "Hctz" is a diuretic drug and is associated to "hydrochlorothiazide" (actually a synonym). Clearly, not all inferred semantic relations are of the same quality.

**Generation of a list of potential biomarkers and phenotypes**: most of the reference biomarkers and phenotypes found in the literature (see Step 1) were linked to one of the following nodes of the ILD: *genes, proteins, causal agents, hormones, phenotypes, metabolic disorders, diabetes, obesity, symptoms*.

The initial pool of candidates was constructed by considering each term appearing in the experimental document base that points to one of the same taxonomy nodes, whether via explicit hypernym relations in the ILD or via inferred hypernyms. For each of these candidates a group of experimental documents was formed by choosing those documents that contain a synonym of the candidate together with synonyms of "diabetes" or "obesity" and for each of these groups the InfoCodex feature vector in semantic space was constructed.

The document group corresponding to one particular initial candidate is compared with the previously derived reference models for D&O biomarkers/phenotypes by computing the semantic distances to the feature vectors of the reference models. A term qualifies as a final candidate for a D&O biomarker or phenotype if the semantic similarity deviation from at least one of the corresponding reference clusters is below a certain threshold.

**Establishing confidence levels:** not all the biomarker/phenotype candidates established this way have the same probability of being relevant. In order to rank the final candidates established in Step 3 an empirical score was devised, representing the confidence level of each term. This confidence measure is based on the average semantic deviation of the feature vector assigned to the candidate from the feature vector of the corresponding reference model and additional information-theoretic measures.

## Results of the experiment

The output of the experiment was a list of potential D&O biomarkers/phenotypes as shown in Table 2. The candidate terms are listed in column A, with their relation to either diabetes or obesity in columns B and C. Columns D and E display the confidence level and the number of documents on which the identification of the candidate was based. Finally, the last columns contain the detailed IDs to these documents so that they can be retrieved and used by human experts for assessment. Note that human expert assessment is actually the only meaningful evaluation of the experiment as far as the novelty aspect of the proposed D&O biomarkers/phenotypes is concerned.

**Table 2: typical output of the experiment**

| Row | Term (A) | Relationship (B) | Object (C) | Conf% (D) | #Docs (E) | PMIDs (F) |
|---|---|---|---|---|---|---|
| 1 | glycemic control | BiomarkerFor | Diabetes | 70.3 | 1122 | 20110333, 20128112, 20149122, |
| 2 | Insulin | PhenoTypeOf | Diabetes | 68.3 | 5000 | 19995096, 20017431, 20043582, |
| 3 | Proinsulin | BiomarkerFor | Diabetes | 67.8 | 105 | 16108846, 9405904, 20139232, |
| 4 | TNF alpha inhibitor | PhenoTypeOf | Diabetes | 67.1 | 245 | 9506740, 20025835, 20059414, |
| 5 | anhydroglucitol | BiomarkerFor | Diabetes | 67.1 | 10 | 20424541, 20709052, 21357907, |
| 6 | linoleic acid | BiomarkerFor | Diabetes | 67.1 | 61 | 20861175, 20846914, 15284064, |
| 7 | palmitic acid | BiomarkerFor | Diabetes | 67.1 | 24 | 20861175, 20846914, 21437903, |
| 8 | pentosidine | BiomarkerFor | Diabetes | 67.1 | 13 | 21447665, 21146883, 17898696, |
| 9 | uric acid | BiomarkerFor | Obesity | 66.8 | 433 | 10726195, 19428063, 10904462, |
| 10 | proatrial natriuretic peptide | BiomarkerFor | Obesity | 66.6 | 4 | 14769680, 18931036, 17351376, |
| 11 | ALT values | BiomarkerFor | Diabetes | 66.3 | 2 | 20880180, 19010326 |
| 12 | adrenomedullin | BiomarkerFor | Diabetes | 64.3 | 7 | 21075100, 21408188, 20124980, |
| 13 | fructosamin | BiomarkerFor | Diabetes | 64.2 | 59 | 20424541, 21054539, 18688079, |
| 14 | TNF alpha inhibitor | BiomarkerFor | Diabetes | 62.1 | 245 | 9506740, 20025835, 20059414, |
| 15 | uric acid | BiomarkerFor | Diabetes | 61.8 | 259 | 21431449, 20002472, 20413437, |
| 16 | monoclonal antibody | BiomarkerFor | Obesity | 61.7 | 41 | 14715842, 21136440, 21042773, |
| 17 | Insulin level QTL | PhenoTypeOf | Obesity | 61.2 | 1167 | 16614055, 19393079, 11093286, |
| 18 | stimulant | BiomarkerFor | Obesity | 61.2 | 646 | 18407040, 18772043, 10082070, |
| 19 | IL-10 | BiomarkerFor | Obesity | 60.9 | 120 | 19798061, 19696761, 20190550, |
| 20 | central obesity | PhenoTypeOf | Diabetes | 59.5 | 530 | 16099342, 17141913, 15942464, |
| 21 | lipid | BiomarkerFor | Obesity | 59.5 | 4279 | 11596664, 12059988, 12379160, |
| 22 | urine albumin screening | BiomarkerFor | Diabetes | 59.0 | 95 | 20886205, 19285607, 20299482, |
| 23 | tyrosine kinase inhibitor | BiomarkerFor | Obesity | 58.8 | 83 | 18814184, 9538268, 15235125, |
| 24 | TNF alpha inhibitor | BiomarkerFor | Obesity | 58.0 | 785 | 20143002, 20173393, 10227565, |
| 25 | fas | BiomarkerFor | Obesity | 57.7 | 179 | 12716789, 17925465, 19301503, |
| 26 | leptin | PhenoTypeOf | Diabetes | 57.6 | 870 | 11987032, 17372717, 18414479, |
| 27 | ALT values | BiomarkerFor | Obesity | 57.4 | 8 | 16408483, 19010326, 17255837, |
| 28 | lipase | BiomarkerFor | Obesity | 56.8 | 356 | 16752181, 17609260, 20512427, |
| 29 | insulin resistance | PhenoTypeOf | Obesity | 55.8 | 5000 | 20452774, 20816595, 21114489, |
| 30 | chronic inflammation | PhenoTypeOf | Diabetes | 55.7 | 154 | 15643475, 18673007, 18801863, |

The details of the evaluation have been published elsewhere [13] and are beyond the scope of the present review. Here I would like to retain the two major conclusions that can be drawn. The negative aspect of the experiment is that too much noise was generated as exemplified by the obviously implausible or incomplete candidates proposed in Table 3.

**Table 3: implausible and/or incomplete D&O biomarker/phenotype candidates**

| Term | Relationship | Object | Target | Conf% | #Docs |
|---|---|---|---|---|---|
| wenqing | BiomarkerFor | Obesity | Obesity | 53.5 | 29 |
| proteomic | BiomarkerFor | Obesity | Obesity | 40.8 | 128 |
| gene expression | BiomarkerFor | Obesity | Obesity | 38.9 | 62 |
| Mouse model | BiomarkerFor | Obesity | Obesity | 19.8 | 17 |
| muise | BiomarkerFor | Obesity | Obesity | 17.5 | 20 |
| athero- | BiomarkerFor | Obesity | Obesity | 16.5 | 6 |
| shrna | BiomarkerFor | Obesity | Obesity | 9.6 | 4 |
| inflammation | BiomarkerFor | Obesity | Obesity | 8.2 | 4 |
| TBD | BiomarkerFor | Obesity | Obesity | 7.4 | 3 |
| body weight | PhenoTypeOf | Diabetes | MGAT2 | | 1 |
| cell line | BiomarkerFor | Diabetes | MGAT2 | | 1 |

The very positive result, however, is that several candidates of **very high quality** were proposed by the software. These were considered as "needles in the haystack" by the Merck experts. While the plausibility of these candidates has been judged very high by human experts, a Google search of these terms in conjunction with "diabetes" and/or "obesity" produced extremely low hit rates, near or at zero, compared with hundreds of thousands for known D&O biomarkers/phenotypes. Unfortunately, these terms are considered as valuable proprietary information by Merck and cannot be shown openly, Table 4.

**Table 4: plausible, novel and very valuable D&O biomarker/phenotype candidates (hidden since considered valuable proprietary information by Merck)**

| *Term* | *Relat.* | *Object* | *Target* | *Conf%* | *#Docs* |
| --- | --- | --- | --- | --- | --- |
| xxxxxx | PhenoTypeOf | Obesity | Obesity | 7.7 | 4 |
| xxxxxx | PhenoTypeOf | Obesity | Obesity | 7 | 6 |
| xxxxxx | BiomarkerFor | Obesity | Obesity | 4.9 | 1 |
| xxxxxx | BiomarkerFor | Obesity | Obesity | 4.9 | 1 |
| xxxxxx | BiomarkerFor | Obesity | Obesity | 2.9 | 2 |
| xxxxxx | BiomarkerFor | Obesity | Obesity | 2.2 | 1 |
| xxxxxx | BiomarkerFor | Obesity | Obesity | 2.2 | 1 |
| xxxxxx | BiomarkerFor | Obesity | Obesity | 2.2 | 1 |
| xxxxxx | BiomarkerFor | Diabetes | Diabetes | 14.5 | 1 |
| xxxxxx | BiomarkerFor | Diabetes | Diabetes | 2.8 | 2 |

Compared with recent studies [28-31] aimed at the extraction of drug–gene relations from the pharmacogenomic literature, this experiment introduced three novelties. First, while most related work is based on high-quality, manually curated knowledge bases such as PharmGKB [30] to train the recognition of connections between specific drugs and genes, this experiment's reference/training set (Step 1) was assembled in an *ad hoc* way by naïve (non-expert) PubMed search. Second, aside from the generic taxonomy in the ILD, no context-specific vocabularies (e.g., UMLS) were provided to inform the semantic engine. The meaning of unrecognized words had to be inferred by the InfoCodex engine based only on its universal internal linguistic database and its association power. Third, the text mining algorithms used here do not use rule-based approaches, or analyze co-occurrences sentence by sentence, or section by section, but rather they extract knowledge from entire documents and their relations with semantically related documents.

In view of the requirement of no human assistance, the demonstrated capability of automatically identifying high quality candidates is extremely encouraging and could prove an entirely new way to speed-up pharmaceutical research, with high potential to shorten time-to-market of novel drugs, or for early recognition of dead ends such as prohibitive side-effects through targeted extraction of relevant information.